\documentclass{article}

\pdfoutput=1
\usepackage{balance}
\usepackage{fullpage}
\usepackage[utf8]{inputenc}
\usepackage{graphicx}
\usepackage{amssymb}
\usepackage{amsmath}
\usepackage{amsthm}
\usepackage{amsfonts}
\usepackage{subfig} 
\usepackage{color}
\usepackage[english]{babel}
\usepackage{algorithm}
\usepackage{algorithmic}
\usepackage{multicol}
\usepackage{multirow}
\usepackage{tabularx,colortbl}
\usepackage{textcomp}
\usepackage{booktabs}
\usepackage{enumerate} 
\usepackage{longtable}
\usepackage{array,longtable,multirow}
\usepackage{lineno}
\usepackage[sort&compress,numbers]{natbib}
\usepackage{authblk}

\definecolor{Orange}{rgb}{1,0.64,0}
\definecolor{lgray}{rgb}{0.9,0.9,0.9}

\makeindex

\title{A Hierarchical Genetic Optimization of a Fuzzy Logic System for Flow Control in Micro Grids}

\author[1]{Enrico {De Santis}\thanks{enrico.desantis@uniroma1.it}\thanks{Corresponding author}}
\author[1]{Antonello Rizzi\thanks{antonello.rizzi@uniroma1.it}}
\author[2]{Alireza Sadeghian\thanks{asadeghi@ryerson.ca}}
\affil[1]{Dept. of Information Engineering, Electronics, and Telecommunications, SAPIENZA University of Rome, Via Eudossiana 18, 00184 Rome, Italy.}
\affil[2]{Dept. of Computer Science, Ryerson University, 350 Victoria Street, Toronto, ON M5B 2K3, Canada.}

\begin{document}

\clearpage \maketitle
\thispagestyle{empty}

\balance


\begin{abstract}
\noindent Bio-inspired algorithms like Genetic Algorithms and Fuzzy Inference Systems (FIS) are nowadays  widely adopted as hybrid techniques in commercial and industrial environment. In this paper we present an interesting application of the fuzzy-GA paradigm to Smart Grids. The main aim consists in performing decision making for power flow management tasks in the proposed microgrid model equipped by renewable sources and an energy storage system, taking into account the economical profit in energy trading with the main-grid. In particular this study focuses on the application of a Hierarchical Genetic Algorithm (HGA) for tuning the Rule Base (RB) of a Fuzzy Inference System (FIS), trying to discover a minimal fuzzy rules set in a Fuzzy Logic Controller (FLC) adopted to perform decision making in the microgrid. The HGA rationale focuses on a particular encoding scheme, based on control genes and parametric genes applied to the optimization of the FIS parameters, allowing to perform a reduction in the structural complexity of the RB. This approach will be referred in the following as fuzzy-HGA. Results are compared with a simpler approach based on a classic fuzzy-GA scheme, where both FIS parameters and rule weights are tuned, while the number of fuzzy rules is fixed in advance. Experiments shows how the fuzzy-HGA approach adopted for the synthesis of the proposed controller outperforms the classic fuzzy-GA scheme, increasing the accounting profit by 67\% in the considered energy trading problem yielding at the same time a simpler RB.\end{abstract}


\textbf{\emph{keywords:}} Microgrid, Energy Management System, Battery Energy Storage, Power Flow Optimization, Storage System Management, Fuzzy Systems, Evolutionary Computation, Hierarchical Genetic Algorithms.


\section{Introduction}
\label{sec:intro}
The world wide power grid can be considered one of  the greatest masterpiece of engineering that human being has ever made. Moreover, starting from the First Industrial Revolution, Smart Grids (SGs) are one of the most important breakthrough that science and engineering fields are carrying out. Currently, the Smart Grid concept is founded on a paradigmatic revolution that will permeate many aspects of human life. From the power system point of view, SGs can be considered as a way to transform the electric energy infrastructures from a centralized, producer-controlled network, into a distributed and consumer-interactive system, leveraging the same concepts and technologies that enabled the emergence and spread of the Internet. \cite{cazzo_minchia_frontier_SG}. SGs vision promises a power grid infrastructure with increased automation of the grid operations and ``self-healing” capabilities. SGs integrate the renewable energy production, seamlessly balancing energy supply and demand, and can be considered as a key technology for facilitating the spread of electric mobility. To achieve that vision, the current power grid has to be thought as a \textit{technological ecosystem} that needs a strong \textit{injection} of distributed intelligence \cite{de2015fault}. 
G. K. Venayagamoorthy argues that the current power grid can be considered a spatially and temporally complex, nonlinear and non-stationary system with a lot of uncertainties \cite{Venayagamoorthy__2009}. Accordingly, the SG can be examined for all intents and purposes as a Complex System, and Computational Intelligence (CI) and Soft Computing (SC) techniques \cite{Ramchurn:2012:PSS:2133806.2133825, 6604117, 5596724, 5952103}, are widely adopted to face a plethora of applications and problems arising in the SG context. The main CI paradigms for SG related problem solutions are: neuro-fuzzy, neuro-swarm, fuzzy-PSO, fuzzy-GA, neuro-Genetic \cite{Venayagamoorthy__2009, 6813686,DeSantis2015, 6846330}. In fact, in MG related tasks, such as control and flow management, the presence of uncertainty and non-linearity, for example in the power demand profile of a large amount of users or in the power produced by solar or wind sources, makes related problems extremely challenging. Hence, SC techniques can help managing the complexity of problems offering reliable solutions, especially in presence of non-linearity \cite{nosratabadi2017comprehensive} and in presence of storage devices that increase the solution space of the unit commitment problem \cite{Fossati201561}.
Consequently, since linear techniques cannot be considered adequate in solving problems whose nature is nonlinear and even stochastic, SC techniques offer a suitable framework introducing learning capabilities in the design of the MG controllers, especially in presence of renewable energy sources and storage.
The current research follows our previous work \cite{de2013genetic} about an application of what we call classic fuzzy-GA paradigm to the problem of Flow Control Optimization in a Microgrid (MG). The MG can be thought as a sub-network of the SG characterized by the presence of autonomous (often renewable) energy sources buffered by some type of Battery Energy Storage System (BESS) and locally controlled in order to achieve smart energy flows management. The flow control task is carried out by a FLC with two Fuzzy Inference Systems (FISs) of Mamdani type. FISs, relaying on approximated reasoning based on Fuzzy Logic (FL), are in charge to optimize energy flows and the overall accounting profit in energy trading operations with the main-grid.
The choice of Mamdani type FIS is related to its simplicity in incorporating human knowledge in the RB. Specifically, in the application at hand the computational overhead introduced by the defuzzification process (absent in a Sugeno type FIS) is not a real problem. Furthermore, the HGA paradigm with its suitable coding scheme is well suited to optimize the structure of a Mandani FIS.
The current work is focused on two main objectives: $i)$ improve the MG model, in particular the BESS model, $ii)$ optimize the Rule Base (RB) of the FLC adopted to control power flows in the MG. As concerns the first goal we move on from an ideal battery adopted in \cite{de2013genetic} designing an energy storage device based on a real-world model with suitable efficiency parameters. For the second objective, we designed an optimization method based on a suited Genetic Algorithm (GA) that is in charge of learning the FIS parameters, optimizing at the same time both the economic return in energy trading and the cardinality of the fuzzy RB. The adopted optimization algorithm is known as Hierarchical Genetic Algorithm (HGA) aiming to perform at the same time a fine tuning of the fuzzy Membership Functions (MFs) and the structural optimization of the FISs -- in the following we will refer on this paradigm as ``fuzzy-HGA paradigm''. 
In fact, while in our previous work the FIS structure is constrained to be fixed, with a predefined number of antecedent and consequent terms and, thereby, with an immutable number of fuzzy rules, the adopted HGA scheme allows to relax these constraints. Moreover, the standard GA approach deals with a chromosome of fixed length whose encoding scheme leads to a lower flexibility in the RB tuning. A GA algorithm able to emulate a variable length chromosome with a suitable encoding scheme of the FIS is ideal for optimizing the number of rules in the given RB. Finally, the fuzzy rule optimization can lead to an improved performance of the FIS, discarding pre-defined low performing rules.
The idea behind HGAs is based on the biological inspired gene structure of a chromosome formed in a hierarchical fashion, emulating the encoding approach of the deoxyribonucleic acid (DNA). In Nature the genes can be classified into two different types: regulatory sequences and structural genes. One of the regulatory sequences found in DNA is called the ``promoter" with the task of activating or deactivating structural genes. Therefore the presence of active and inactive genes in the structural genes leads to the idea of a hierarchical structure formulation of the chromosome that consists of \textit{control genes} and \textit{parametric genes}. The activation of the parametric genes is governed by the value of the control genes. Accordingly, the strategy suggested by Nature can be modeled for solving a number of engineering problems, such as those involving mix integer programming methods \cite{Pierrot97} or fuzzy control applications demanding the joint optimization of both FISs parameters and RBs \cite{Ko06}. In the last case the novelty of a hierarchical coding scheme is based on the definition of suitable genetic operators moving from standard GA algorithms to more advanced ones. The hierarchical encoding scheme allows to code the FISs parameters, more precisely MF parameters, as parametric genes and, at the same time, control genes can be used to activate and deactivate MFs composing a given fuzzy RB, thus tuning the overall number of fuzzy rules.
The work is organized as follows.\\
Sec. \ref{sec:Related_works} is a literature review about the use of FLC and the fuzzy-GA paradigm in the Smart Grid context. In Sec. \ref{sec:Background} we introduce the optimization problem and the MG model. Sec. \ref{sec:PF} clarifies the level of abstraction of the problem, introducing the adopted notation and explaining how the fuzzy controller works. The fuzzy control scheme for a MG, together with the classic fuzzy-GA and fuzzy-HGA paradigms are treated in Sec. \ref{sec:Fuzzy_C_MG} and related subsections.
In Sec. \ref{sec:Experim_ev}, soon after the introduction of the examined MG scenarios and the algorithm settings, the main results are reported and discussed. Finally, conclusions are drawn in Sec. \ref{sec:conclusions}, while feature works are discussed in Sec. \ref{sec:FWS}.


%
%
\section{Related works}
\label{sec:Related_works}
In the SC field, FL is a well consolidated discipline able to integrate approximate reasoning methodologies in engineering systems. SC techniques are commonly adopted in applications dealing with uncertainties \cite{6863314,6504816}. An interesting review on fuzzy logic and its hybrid approaches employed in the Smart Grids and microgrids context can be found in \cite{suganthi2015applications}.
In \cite{6936001} a small-scale MG designed to provide power to local communities, able to connect or disconnect from the main-grid, is studied. The control scheme is based on a non optimized FLC considering as input the electricity prices, the renewable production, and the load demand. The RB is developed in order to ensure reliable grid operations taking into account the financial aspect to decide the load modification's level. FLCs can be designed to control the amount of power that should be taken from the battery system in case of power deficiency to cover the load demand in a scheduling process \cite{nosratabadi2017comprehensive}.
The latter problem is faced in \cite{Mohamed2013133}, optimizing distribution system operations in a SG, from cost and system stability points of view. In the adopted MG model the FLC was used only in the case when the instantaneous load demanded is higher than the instantaneous available power from renewable energy sources and the system is not at the peak period (the battery is operated in discharge mode). However the FLC does not foresee any further optimization.
In \cite{7454775} is exploited a FLC in a MG Energy Management System (EMS) equipped by a BESS in order to apply peak shaving and valley filling and to limit the oscillation of energy exchanged with the network buffering the ESS. The FLC aims to reduce the variation of the energy exchanged with the grid. Moreover, it exploits a low bandwidth filter set before the FLC. The FLC is also optimized, but the adopted algorithm is not specified. 
In \cite{6157610} it has been considered a MG grid-connected equipped with a battery and a fuel cell. The MG consists of a wind turbine, a PV plant and a micro-turbine; the EMS must satisfy the load demand.
In this work, the considered objective functions to be minimized are the energy expenses and the emissions of a MG, through multi-objective optimization.
The proposed EMS approach consists in the joint formulation of multi-objective optimization approach based on linear programming and battery scheduling (working both on-line). The battery scheduling, as a part of an optimal online energy management, is made through a FLC designed by an expert operator for the on-line scheduling of the BESS. It defines whether the battery should be charged or discharged and at which convenient rates. One of the FLC input is the predicted energy in the day ahead both for the load and the generation. After the scheduling of the BESS, the remainder of the power flows are optimized with a multi objective linear programming algorithm.
In the MG modeling field a FL-based framework is proposed in \cite{6320896} aimed at controlling a BESS to achieve an high efficient management. Authors propose a MG model able to operate synchronously with the main-grid (grid connected mode) or independently (islanded mode) aiming to control the amount of power delivered to/taken from the storage unit in order to improve a cost function based on purchasing/selling power to the main-grid. Predetermined human reasoning-based fuzzy rules are adopted to define a FLC with inputs variables such as consumer's load demand, electricity price rate and renewable energy generation rate. Specifically, data are sampled every 15 minutes and the evaluated figure of merit consists of the summation of payments/revenues (similarly to the current work) along with system losses. However, the fuzzy RB is kept fixed, thereby any subsequent learning and optimization procedure is not considered. 
It is well known that FL, especially in FLC of Mamdani type, is well-suited in incorporating human knowledge in artificial systems.
However, the synthesis of the fuzzy rules in a complex control problem can be a challenging task and in presence of many free parameters in the RB the human expertise is not sufficient to assure satisfying performances. Thereby, the fuzzy-GA modeling can help to improve the inherent human approximation level in generating fuzzy rules. Furthermore, an evolutionary strategy can fill the gap between a fixed RB and a flexible one introducing a suitable learning scheme. Thereby, an HGA is able to tackle the problem of managing many free parameters of a Mamdani FIS, optimizing at the same time the number of rules and selecting the ones that perform better in terms of the considered objective function. 
In \cite{Environmentally_Friendly_Control}, the fuzzy-GA approach is adopted for controlling energy flows and for managing the MG, in presence of an electric storage unit and distributed renewable energy generation. The main goal consists in optimizing a cost function defined as the net expense in electric energy purchasing from the main-grid. 
In \cite{Fossati201561} authors propose, within the fuzzy-GA paradigm, an algorithm for optimizing the MG operations. In particular  given the load demand, the wind power generation and day-ahead electricity prices, the algorithm is able to set up a MG generation schedule and to synthesize a suitable fuzzy expert system for allocating the energy from the storage system. Moreover, two GAs are used alternatively, one to determine both the microgrid scheduling and the fuzzy rules, the other to tune the MFs. The proposed expert system, responsible for setting the power withdrawn from the battery, is provided with learning capabilities so that it can optimize its knowledge base (fuzzy rules and MFs) according to the given scenario.
A FL-based energy management, adopting a fixed Mamdani FIS, is studied in \cite{chehri_feman_2013} for house consumption applications in presence of a solar source. The main aim here is maximizing the use of solar energy and reducing the impact on the power grid, while satisfying the energy demand of house appliances, giving a fixed level of comfort.


\section{Background}
\label{sec:Background}
\begin{figure*}[!ht]
 \centering
 \includegraphics[scale=1.2,keepaspectratio=true]{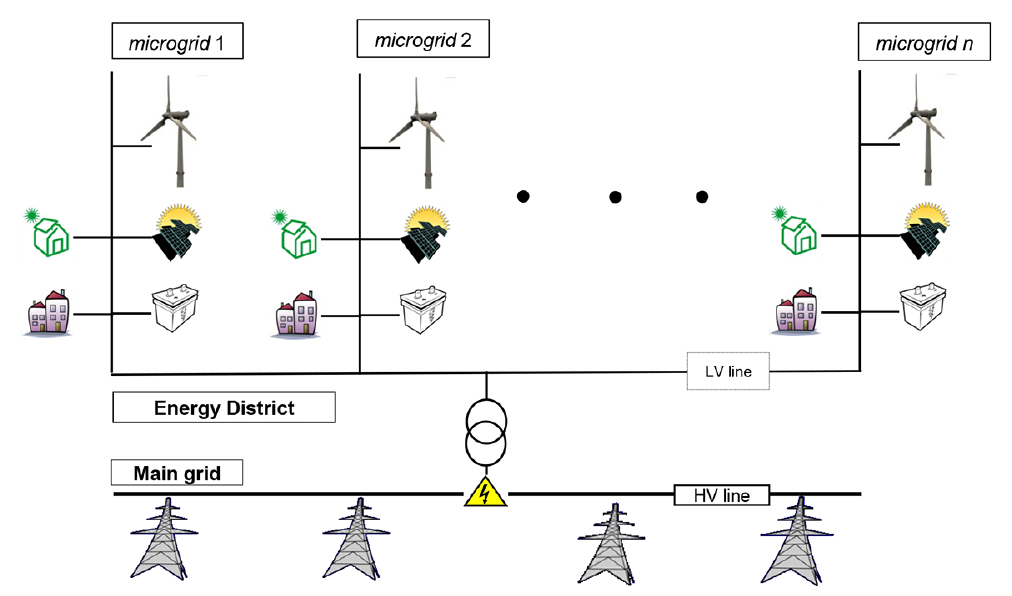}
 \caption{Scheme of a common Energy District.}
 \label{fig:Energy_District}
\end{figure*}

The proposed approach concerns a control scheme relating to energy flows of a MG belonging to an energy district -- see Fig \ref{fig:Energy_District} -- and connected to the main-grid or even other MGs. The MG can perform its operations in a ``grid connected mode" and in ``islanded mode". In the proposed model the energy production is typically provided by renewable Distributed Resources (DRs) such as solar, wind,  and micro-hydro generators. In addition, the MG can be equipped with a traditional energy source, such as a turbogas generator. Moreover, the MG is provided by a Li-ion BESS; the model is compliant with technical specifications of a specific Toshiba device. The BESS is controlled through a Battery Management System (BMS) and can be used to compensate the lack of energy production when DRs are unable to meet domestic demand or to accumulate energy in case of overproduction. The BMS can be programmed also to store energy that can be resold to the main-grid when favorable economic conditions apply (based, for example, on energy price signals).
The main scenarios, in which the proposed controller can act, are: 

\begin{itemize}
\item the energy production within the MG meets (or exceed) the energy demand;
\item the energy production within the MG does not meet the demand.
\end{itemize}
In the former case the MG operates in ``islanded mode" and the BESS can be used to store a fraction of the surplus of energy, selling the remainder to the main-grid. In the latter case the MG can operate in ``grid connected mode" and can satisfy the loads by drawing a fraction of the negative energy balance from the BESS, while buying the remainder from the main-grid.
The two main operations introduced above depend strictly on the MG internal state that includes the State of Charge (SOC) of the BESS and thus the available stored energy, the current price of energy and the generated and demanded power.
 
It is noted that the two main objectives of the proposed controller are:
\begin{itemize}
\item to guarantee the continuous to power supplies of energy availability without interruptions. Thus the MG must guarantee an high quality of service.
\item to maximize the accounting profit for possible energy trading with the main-grid or other MGs,   given the collected information about the energy price and the MG internal state.
\end{itemize}

\begin{figure*}[htp]
 \centering
 \includegraphics[scale=0.50,keepaspectratio=true]{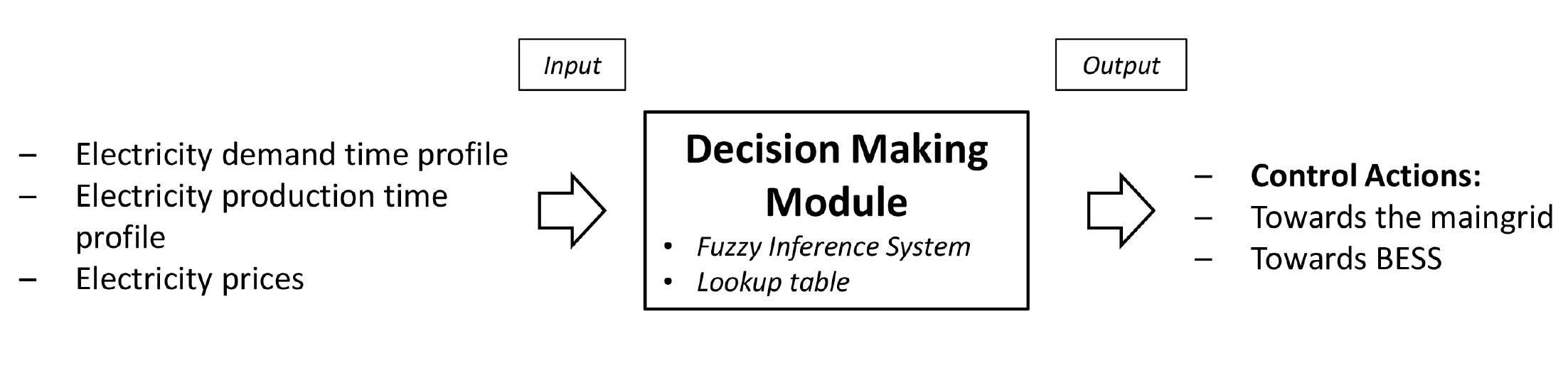}
 \caption{Simplified scheme of the control module for a given MG.}
 \label{fig:S_C_Scheme}
\end{figure*}

In Fig. \ref{fig:S_C_Scheme} is depicted the simplified control scheme for a given MG. The controller is immersed in an external environment where information is assumed to be originated from a centralized data-center or directly from other connected MGs, in an automatic cooperative/competitive trading scenario. Input information to the controller are the aggregated energy time profile requested from the MG loads, the aggregated energy time profile produced by the DRs, the energy price signal coming from the main-grid and the SOC of the BESS. The output of the controller is a pair of signals towards the main-grid or the BESS.
The decision making module adopts a fuzzy inference scheme, implementing two distinct FISs for deciding the best actions trying to maximize the accounting profit of the MG in a given time span. A Hierarchical Genetic Algorithm is adopted to tune the set of fuzzy rules and their parameters and thus to learn their best combination exploiting a given suitable objective function.
\begin{figure*}[htp]
 \centering
 \includegraphics[scale=1.2,keepaspectratio=true]{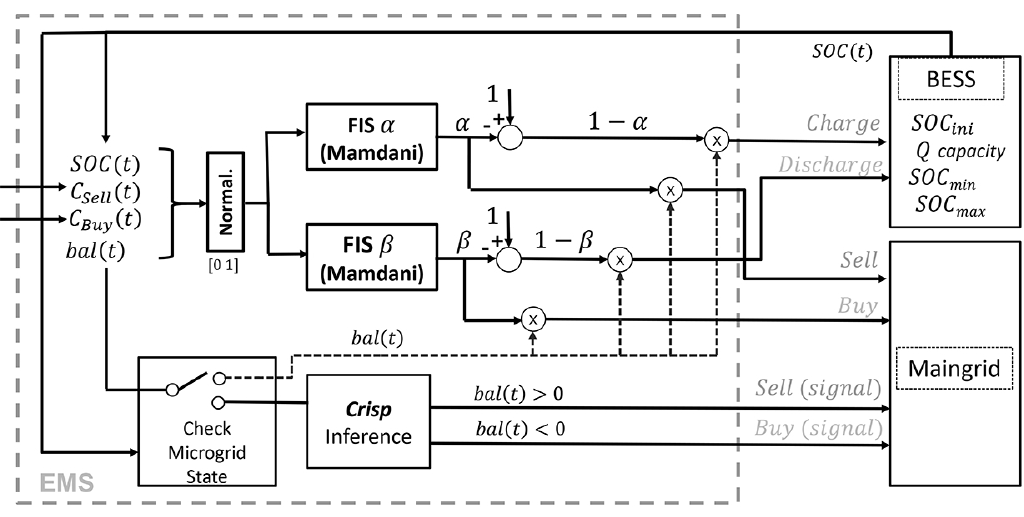}
 \caption{Functional diagram of the controller.}
 \label{fig:FD}
\end{figure*}

\section{Problem Formulation}
\label{sec:PF}

\subsection{Model's assumptions}
\label{sec:Model}
The entire control chain in a general microgrid can be considered as organized as a three level system: the field level (data acquisition and decentralized control), the SCADA level (supervision and control functions), and the planning and management level, which can be a decentralized EMS \cite{nosratabadi2017comprehensive}. In this work, we focus on the third level. Thereby, the model that we will discuss is based on a number of hypothesis that define the level of abstraction useful to correctly place the problem under analysis.
The main assumptions are reported below.
\begin{itemize}
  \item The considered timestamps are discretized and monospaced.
  \item The present study does not consider low level operations such as voltage and reactive power control. 
  \item Only high level operations are evaluated, such as the control of energy flow within a MG.
  \item In the simulation several energy prosumers are contemplated, each one has its own characteristic energy profile with a timing defined each quarter of an hour.
  \item For each time sample the flow control system takes into account the aggregate energy production and demand.
  \item In each time sample the control system is able to fulfill a decision making request.
  \item The demanded energy is actually consumed.
  \item The main-grid has an unlimited energy capacity.
  \item The current version of the MG flow control system deals with a single BESS, as an overall model of the total storage capacity within the MG (possibly due to more than one distinct devices).

  \item The current work does not cover data exchange models within the main-grid and it is supposed that the MG is equipped with a suitable data network infrastructure with low delay and high QoS in exchanging data packets.
\end{itemize}
Hence, the adopted controller focus on the energy exchanges between portions of the MG and the main-grid considered fairly of unlimited capacity given the powers involved in small-scale MGs, as the one herein considered. Furthermore a discrete data model for the measured power consumed and demanded is used with a sample time of 15 minutes, as adopted in most high level control schemes. 
The last assumption, about the low delay in data exchange, is reliable considering both the low delays characterizing communication protocols used in modern SGs and the small extension of the adopted MG model. For instance, cellular wireless networks such as GPRS, UMTS, or 4G technologies like 802.16m and LTE could be used for the interface between smart meters and the central system \cite{6102299,6102301}. Furthermore, multiple communication technologies and standards could coexist in different parts of the Smart Grid, especially in a multi-microgrid context, and their impact deserves a deep investigation of communication devices and channel networks.

\subsection{Notation}
\label{sec:Notation}
In general, given a MG like the one depicted in Fig. \ref{fig:Energy_District}, thus consisting of several energy generators and loads, the aggregate electricity production can be defined as:
\begin{equation}
P_{agg}(t)=\sum_{k=1}^{K}P_{k}(t),
\label{eq:Prod_agg}
\end{equation}
where $P_{k}(t)$ is the electrical power produced by the $k$-the producer within a set of $K$ producers, at time sample $t$.\\
Similarly, we can define the aggregate electrical power demand within the given MG:
\begin{equation}
D_{agg}(t)=\sum_{l=1}^{L}D_{l}(t),
\label{eq:Cons_agg}
\end{equation}
where $D_{l}(t)$ is the amount of the demanded electricity effectively consumed by the $i$-th user within a set of $L$ consumers, at time sample $t$.
The MG is equipped with a BESS, whose detailed model will be provided in Sec. \ref{sec:BM} below. Hence, at the abstraction level of the controller, the electricity storage can be characterized by the SOC variable, defined as: 
\begin{equation}
\label{eq:SOC-tau}
SOC_{min} \leq SOC(t) \leq SOC_{max},\quad SOC_{min} \geq 0,
\end{equation}
where $SOC_{min}$ and $SOC_{max}$ represent the lower and the maximum threshold values for the SOC, respectively. In other words, these values constitute the interval of admissibility for the charge level of the battery unit.\\
Considering both the aggregate demand time profile (\ref{eq:Prod_agg}) and the aggregate production time profile (\ref{eq:Cons_agg}) within the given MG, the power balance can be defined as:
\begin{equation}
Bal(t)= P_{agg}(t) - D_{agg}(t);
\label{eq:bal}
\end{equation}
Finally, we will denote with $C_{buy}(t)$ and $C_{sell}(t)$ the price signals in buying and selling energy from/to the main-grid, respectively.

\subsection{MG State}
\label{sec:MG States}
The MG state at time sample $t$, from the point of view of the controller, depends on the energy balance, the energy price signals and the State Of Charge (SOC) of the storage unit. Therefore, it is possible to define a generic MG state as:
\begin{equation}
S(t)= \{ Bal(t), SOC(t), C_{buy}(t), C_{sell}(t) \}.
\label{eq:states}
\end{equation}
The energy balance value (\ref{eq:bal}) defines three distinct sub-states in which the MG can be found, namely:
\begin{enumerate}
\item if $Bal(t)<0$, the MG has a deficit of energy (underproduction state);
\item if $Bal(t)=0$, the produced energy fulfills the internal demand (equilibrium state);
\item if $Bal(t)>0$, the MG has a surplus of energy (overproduction state).
\end{enumerate}
For each given sub-state we can have the following contingencies for the storage unit:
\begin{enumerate}
\item If $SOC(t)=SOC_{min}$, the available charge has a null value;
\item If $SOC(t)=SOC_{max}$, the charge has a maximum value;
\item If $SOC_{min} \leq SOC(t) \leq SOC_{max}$, the available charge has an intermediate value.
\end{enumerate}
The above reported fine grained classification leads to define a set of control actions taking into account also the energy price signals.

\subsection{Control Actions}
\label{sec:CA}
The inference process underlying the controller is performed by a set of control actions dispatched to actuators. The control actions depend on the MG state $S(t)$. 
It is noted that, in the following, the term ``\textit{crisp}" \cite{klir1995fuzzy} relies to the distinction made in FL among \textit{crisp} sets, that are ordinary sets and fuzzy sets defined by a MF, as it will discussed in the following Sec. \ref{sec:Fuzzy_C_MG}.  
The set of control actions is defined as:   
\begin{equation}
\begin{aligned}
\mathbf{a}^{tot}= \{& FORCED\_CHARGE,\;   \\        
                  & FORCED\_BUY,\;  \\
                  & FORCED\_SELL,\; \\
									& (1-\alpha) \, CHARGE \wedge \alpha \, SELL,\; \\
									& (1-\beta) \, DISCHARGE \wedge \beta \, BUY \} 
\end{aligned}
\end{equation}
Hence, for the generic action we have:
\begin{equation}
\mathbf{a}(t) \subseteq \mathbf{a}^{tot}.
\end{equation}
In addition, the set of actions just defined can be split in two disjoint subsets:
%
%
\begin{enumerate}
\item ``forced" action that are ``crisp" action due to: $i)$ the energy deficit in the MG and the lack of available BESS charge, or $ii)$ the energy surplus in the MG and the capacity limit reached by the BESS;
\item ``fuzzy" actions that depend on two real valued parameters $\alpha, \beta \in [0,1]$ encoding respectively: 
\begin{enumerate}
\item the quantity of energy to sell to the main-grid or used to feed the battery;
\item the quantity of energy to purchase from the main-grid or to draw from the battery.
\end{enumerate}
\end{enumerate}
The disambiguation task about these cases is performed by a suitable module -- see Fig. \ref{fig:FD} -- able to evaluate the MG  state $S(t)$, in charge of selecting the activation of the crisp decision making procedure or the fuzzy one.\\
For the first subset of control actions it is defined the energy exchange in input and output from/to the main-grid, as: 
\begin{eqnarray}
\label{eq:Pin}
E_{out}(t)=\alpha(t) \, Bal(t)= \alpha(t) \, [P_{agg}(t)-D_{agg}(t)] \Delta T  \\
E_{in}(t)= \beta(t) \, Bal(t)= \beta(t) \, [P_{agg}(t)-D_{agg}(t)] \Delta T ,
\label{eq:Pout}
\end{eqnarray}
where $\alpha$, $\beta$ are real numbers in $[0,1]$ and $\Delta T$ is the time slot width.
It is noted that the unitary value for both the $\alpha$ and $\beta$ control parameters leads to actions that overlaps with the set of ``crisp" actions. However, these ones must be classified as \textit{deliberate} choice leaded by the inference module, rather than ``forced actions".
The second subset of control actions refers to the energy exchanges with the energy storage device, that at the abstraction level of the controller, is governed with the following dynamical equation regarding the SOC \cite{6033031}: 
\begin{equation}
SOC(t)=SOC(t-1)+h(t),
\label{eq:SOC_eq}
\end{equation}
where $h(t)$ is the signed quantity of charge exchanged with the MG, as required at each time step by the FLC.

\subsection{Market Policies and Objective Function}
\label{sec:FPOF}
Given the purchasing and selling price signals and the energy exchanged with the main grid \ref{eq:Pout}, it follows that the economical balance of the MG is: 
\begin{equation}
Profit(t)=Revenue(t)-Expense(t).
\label{eq:Profit}
\end{equation}
The last equation is the accounting profit (or the accounting loss, depending on the sign of the first term of (\ref{eq:Profit}) for each time sample and it is the difference between the revenues and the expenses arising from the energy trading. 
In Eq. (\ref{eq:Profit}) the revenues and the expenses are defined respectively:
\begin{equation}
Revenue(t)= C_{sell}(t) \, P_{out}(t) \Delta T  \\
\label{eq:rev_exp}
\end{equation}
\begin{equation}
Expense(t)= C_{buy}(t) \, P_{in}(t) \Delta T, 
\label{eq:rev_exp2}
\end{equation}
where $C_{sell}(t)$ is the selling price and $C_{buy}(t)$ is the purchasing price. The last two equation in (\ref{eq:rev_exp}),(\ref{eq:rev_exp2}) and thus the (\ref{eq:Profit}) are functions of the parameters $\alpha$ and $\beta$ that are the output of the inference process based on the FLC. The presence of the energy storage capacity that constitutes a degree of freedom of the control system allows, for example, in the case of surplus of the energy production to store energy if the current selling price is low and to sell the stored energy when the price is high. On the basis of the economic policies chosen for the given MG, it is possible to define an objective function that the controller has to fulfill. The objective function introduced in the current work, that is optimized by an evolutionary algorithm capable of fine tuning the FLC parameters, is the overall profit on a finite temporal horizon of $N$ time samples. Therefore, the objective function is defined as:   
%
\begin{equation}
\begin{split}
&\sum_{t=1}^{N}Profit(t)=\sum_{t=1}^{N}Revenue(t)-Expense(t).\\
\label{eq:Obj_f}
\end{split}
\end{equation}
\subsection{The Battery Model}
\label{sec:BM}
\begin{figure}[ht!] 
\centering
\includegraphics[scale=0.50]{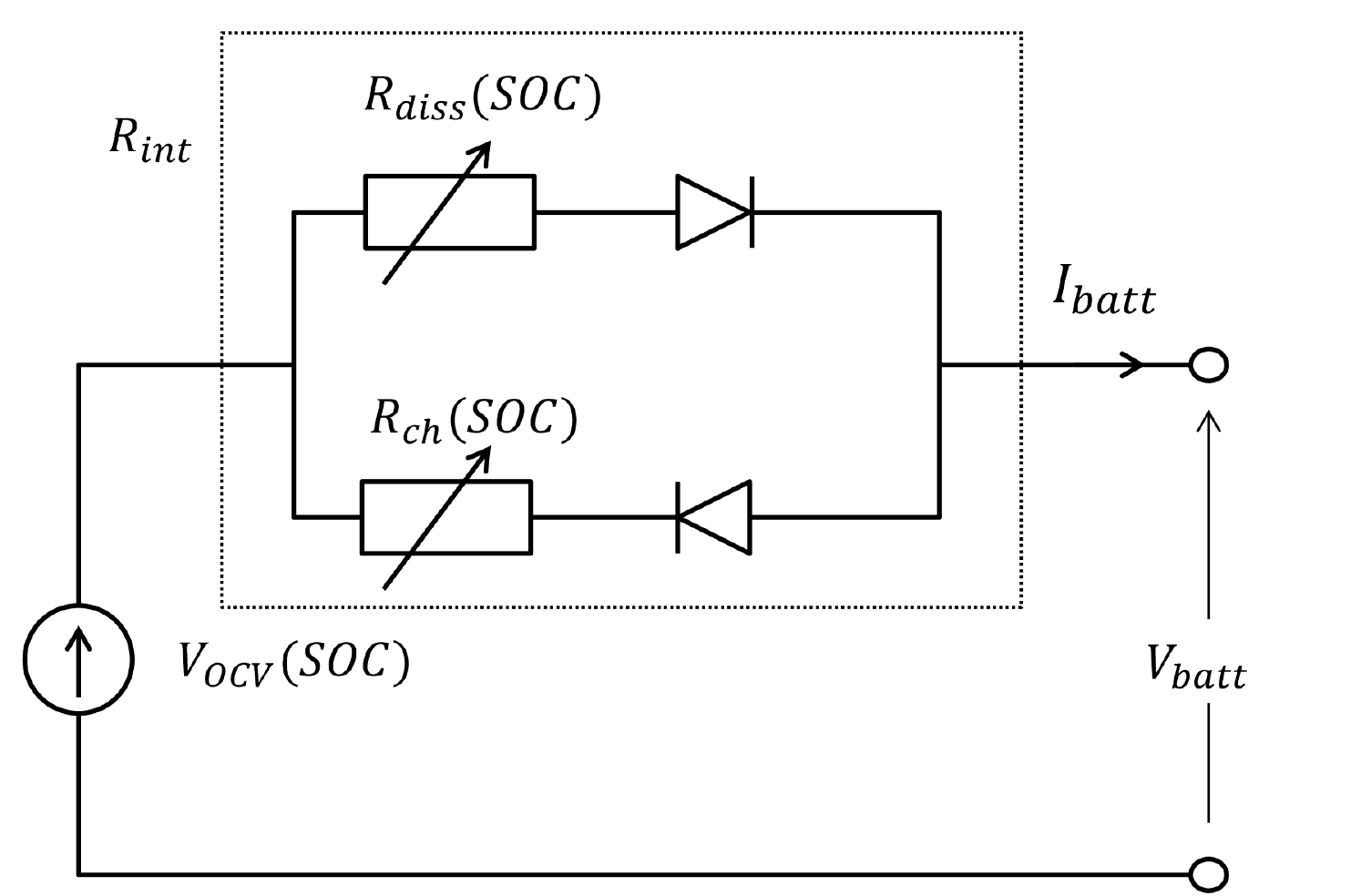}
\caption{ The battery equivalent circuit.}
\label{fig:batt_scheme}
\end{figure}

At the abstraction level of the controller the energy storage capacity linked to the given MG is modeled by means of the (\ref{eq:SOC_eq}), through the temporal dynamic of the SOC parameter \cite{6033031,de2013genetic}. In order to exploit a more realistic model of the MG, the BESS model is enriched with physical real-world characteristics regarding battery devices currently used in battery-to-grid applications. Precisely, the model proposed in the current section is based on a real-word Li-on battery suitable for grid-tied applications \cite{doi:10.1021/cr100290v,6033031,6816050,Luo2015511}.  Today, Li-on batteries are not only used in consumer electronics but also in Smart Grid, as energy storage capacity for MGs or in V2G applications \cite{6623096,Li20103338,Li2009542}. The Li-on technology is characterized by both a good ratio power/weight (about 150 Wh/Kg) and an high efficiency that exceed 90\%. A fundamental parameter in a battery model is the efficiency $\eta_{e}$, defined as the energy fraction that we can recover from the battery after storing a given amount of energy. The efficiency parameter can be represented as:
%
\begin{align}
&\eta_{e}=\frac{\int_{0}^{t_{s}}I_{s}(t)\;V_{s}(t)\;dt}{\int_{0}^{t_{c}}I_{c}(t)\;V_{c}(t)\;dt}\;,\; \\
&\mbox{with:}\;\Bigg\lbrace 
\begin{array}{ll}
V_{s},\,I_{s}=\mbox{Discharging Current and Voltage} \nonumber\\
V_{c},\,I_{c}=\mbox{Charging Current and Voltage}
\end{array}
\end{align}
%
Moreover, starting from the equivalent circuit of a generic battery -- see Fig. \ref{fig:batt_scheme} -- it is possible to outline the fundamental parameters able to increase the accuracy of the battery model. The model foresees an ideal voltage source $V_{OCV}$ that consists in the open-circuit voltage of the battery and an internal resistance $R_{int}$. The $I_{bat}$ and $V_{bat}$ represent respectively the output current and voltage. From the circuit scheme depicted in Fig. \ref{fig:batt_scheme}, $V_{OCV}$ and $R_{int}$ depend on the SOC value, hence:
\begin{align}
&V_{OCV}=f_{1}(SOC),\\
&R_{int}=
  \left\{\begin{aligned}
         R_{ch} & =f_{2}(SOC) & \mbox{Charge}, \\
		 R_{dis} & =f_{3}(SOC) & \mbox{Discharge}.
        \end{aligned}
  \right.
\end{align}

\begin{align} 
&V_{bat} = V_{OCV} - R_{int} \, I_{bat} \label{formula-vocv},\\ 
&SOC = SOC_{ini} - \int_{0}^{t_{s}} \frac{\eta \, I_{bat}}{Q} dt \label{formula-soc},\\
&\eta=
  \left\{\begin{aligned} \label{formule-eta}
         \eta_{ch} & =\frac{V_{OCV}}{V_{OCV}-R_{ch}\,I_{bat}} & \mbox{Charge}, \\
		 \eta_{dis} & =\frac{V_{OCV}-R_{dis}\,I_{bat}}{V_{OCV}} & \mbox{Discharge}.
         \end{aligned}
  \right.
\end{align}

The $\eta_{ch}$ and the $\eta_{dis}$ values are respectively the charging and discharging efficiency, while $Q$ is the capacity of the battery (it is measured in Ah) representing the amount of electric current intensity that the battery is able to supply in a time interval of one hour. Both the efficiencies $\eta_{ch}$ and $\eta_{dis}$ together with the internal resistance $R_{int}$  can take values depending on the state of the battery. Therefore, $V_{OCV}$ and $R_{int}$ are variables that are functions of the charging state \cite{5485606}.
%
%
\section{Fuzzy control of a MG}
\label{sec:Fuzzy_C_MG}
FL is a very useful paradigm in dealing with fuzzy concepts expressed by fuzzy words (i.e. High, Low, Warm, Cold, etc.) in computational and algorithmic frameworks and, as Lotfi A. Zadeh stated: Fuzzy Logic means ``computing with words" \cite{493904}. 
Moving from this general concepts, in Engineering and in particular in the Control Systems field, researchers developed the Fuzzy Logic Modeling (FLM) paradigm to tackle control problems in complex systems aiming to model the underlying uncertainty \cite{52551}. As baseline for the control scheme of the proposed MG model, in our previous work \cite{de2013genetic} it is developed a FLC of Mamdani type capable to express an approximate reasoning method in order to control a real-world system, integrating knowledge coming from human experts \cite{MAMDANI:1999:ELS:330379.330383}.\\
\begin{figure}[htbph]
\centering
\hspace{-0.2cm}
\includegraphics[scale=0.7]{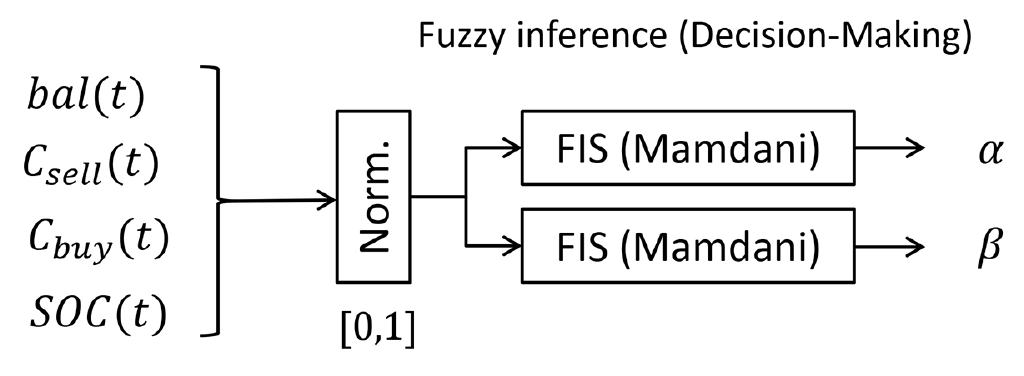}
\caption{Functional diagram of the decision making module.}
\label{fig::func_diag_dec_mak}      
\end{figure}%

\subsection{The basic Fuzzy Control Scheme}
\label{sec:SFCS}
As established in Sec. \ref{sec:CA}, the controller can manage a particular type of control rules that depend on two real-valued parameters, namely $\alpha$ and $\beta$. Such parameters are currently computed by means of two FISs whose scheme is depicted in Fig. \ref{fig::func_diag_dec_mak}. Formally, input variables, normalized in the real-valued range [0,1], are computed separately by two functions ${\Phi _1}$ and ${\Phi _2}$ defined as follows: 
\begin{subequations}
\begin{align}
\alpha (t) = {\Phi _1}(bal(t),SOC(t),{C_{sell}}(t)) \\
\beta (t) = {\Phi _2}(bal(t),SOC(t),{C_{buy}}(t))
\end{align} 
\label{eq::param}
\end{subequations}
Both FISs are based on a Mamdani type fuzzy linguistic model able to introduce semi-qualitative if-than fuzzy rules. The paradigm used for our problem is the Multiple Input Single Output (MISO). Hence, given a fixed number of inputs, a MISO FIS compute a single real-valued output. More specifically, given $p$ antecedents in a conjunctive form, the structure of the fuzzy rule can be written as:
\begin{equation}
\begin{split}
{R_i}{\rm{:}}\quad {\rm{If}}\;{x_1}\;{\rm{is}}\;{A_{i1}}\,{\rm{and}}\,{x_2}\,{\rm{is}}\,{A_{i2}}\,{\rm{and}}...{\rm{and}}\,{x_p}\,{\rm{is}}\,{A_{ip}} \\
{\rm{then}} \;y\;is\;{B_i}\,\,{\rm{with}}\,{w_i},\quad {w_i} \in [0,1],\quad i = 1,2,...,K.
\end{split}
\label{eq::fuzzy_r}
\end{equation}
where $x_j$ is a real number interpreted as a singleton MF, $A_{ij}$ and $B_i$ are fuzzy labeled sets, i.e., the antecedent and consequent linguistic terms, and $w_i$ are rule weights measuring the $i$-th rule strength among a set of $K$ fuzzy rules.
Moreover, the \textit{Fuzzy Conjunction} of the antecedents  i.e., the ``and" operator, and the \textit{Fuzzy Implication}, i.e. the ``then" operator, are suitable \textit{triangular norm} \cite{52551,52552}.
Fuzzy rules, representing fuzzy implications or relations  \cite{52551}, in a conjunctive form decomposes the input domain into a set of fuzzy hyper-boxes, obtaining a \textit{grid partition} of the FIS input domain. If FIS parameters are tuned by an optimization procedure the partition is said to be adaptive \cite{Jang:1996:NSC:248321}. Nowadays, numerous fuzzy reasoning models have been developed and they are characterized by the number and the type of parameters, the shape of MFs and the ``defuzzification methods". The last refers to a procedure to obtain a single scalar value from a fuzzy set \cite{Leekwijck1999159}. 

Within the fuzzy-GA paradigm and in the simpler learning scheme, a Genetic Algorithm (GA) is adopted to optimize only the MF parameters. The rationale is that the GA evolves a set of FISs and returns, by means of an evolutionary procedure imitating the natural selection, the best among a population in terms of a suitable and well-defined performance measure or objective function.
\subsection{Fuzzy Logic Controller optimization: the classic scheme}
\label{sec:Simple_scheme}
\begin{figure}[htbph]
\centering
\hspace{-0.2cm}
\includegraphics[scale=0.55]{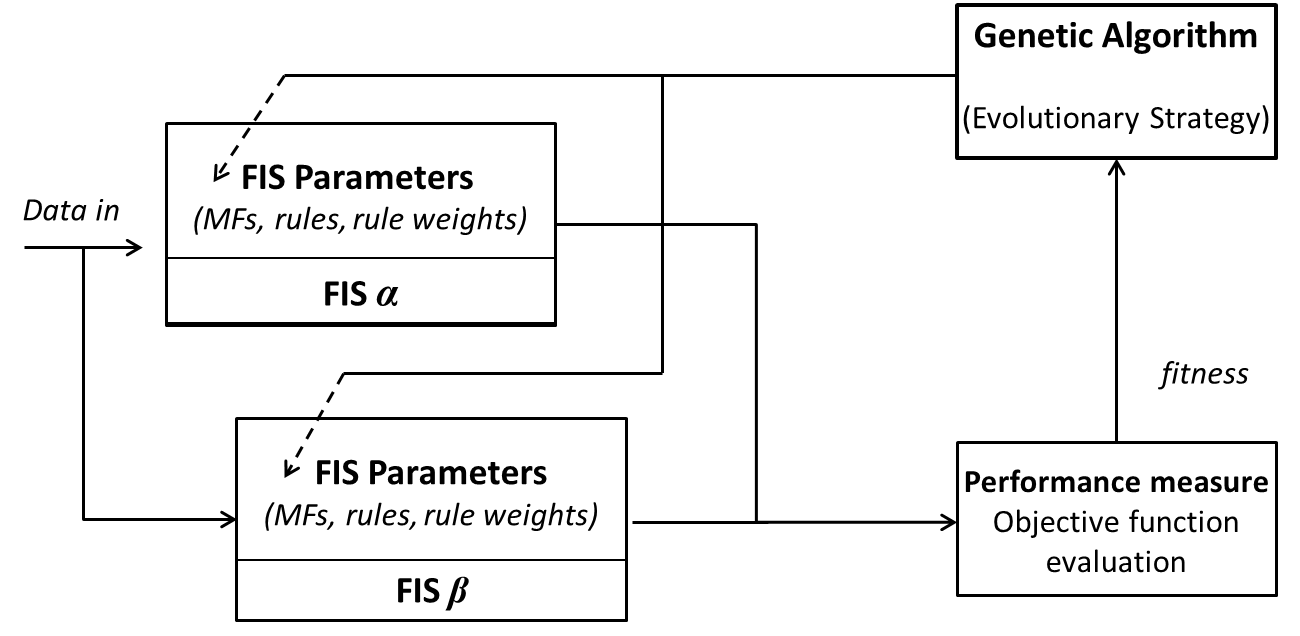}
\caption{System architecture for the fuzzy-GA paradigm for the Flow Control model.}
\label{fig::gen_opt}      
\end{figure}

\begin{figure}[htbph]
\centering
\includegraphics[scale=0.57]{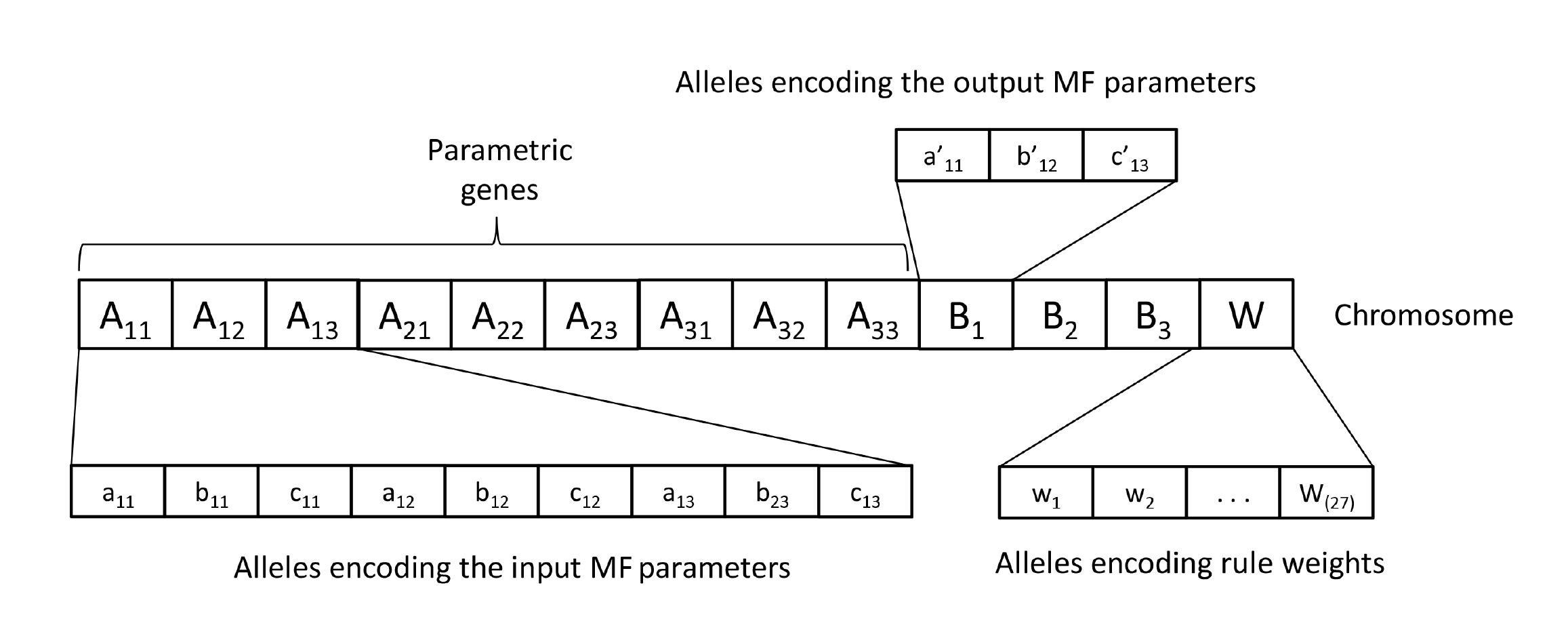}
\caption{Chromosome structure: genes that encode for the MF parameters and the weights of fuzzy rules (\textit{descriptive} fuzzy reasoning method).}
\label{fig::enc_scheme}      
\end{figure}

Depending on the parameters of a FLC that one decides to tune, we can fulfill different parameter learning schemes. In the first fuzzy control scheme, that is the simpler, the GA is applied to a procedure for \textit{knowledge learning} and in part of \textit{structure learning} \cite{Wagholikar}. The basics of the optimization scheme for the proposed control problem, obtained through a standard GA is depicted in Fig. \ref{fig::gen_opt}. For each of the three input in (\ref{eq::param}) and for the output we employed three fuzzy labels, while for the shape of the input and output MFs we adopted the triangular ones.
The form of a general fuzzy rule with three antecedents is:
\begin{equation}
\begin{split}
{\rm{IF}}\;{x_1}\; &{\rm{IS}}\;{A_{i1}}\;{\rm{AND}}\;{x_2}\;{\rm{IS}}\;{A_{i2}}\;\;{\rm{AND}}\;{x_3}\;{\rm{IS}}\;{A_{i3}}\, \\ &{\rm{THEN}}\;y\;{\rm{IS}}\;{B_i}\;with\;{w_i}.
\end{split}
\label{eq::fuzzy_r2}
\end{equation}
For the number of fuzzy rules we have chosen a maximal set, that is the maximum allowed by a regular grid partition, induced by the fuzzy rules structure defined in \eqref{eq::fuzzy_r2}, namely 27.
Having three inputs and working with triangular MFs the \textit{descriptive} fuzzy reasoning method leads to a simple encoding scheme --  see Fig. \ref{fig::enc_scheme} -- for the FLC parameters, namely the MF parameters and the rule weights $w_i$. For both the FISs each gene represents a particular FIS instance and each allele represents the parameter subject to the evolutionary tuning. Having three parameters for each triangular MF with three input and one output for the FLC, the length of the overall chromosome accounts for 
$9+9+9+9 = 36$, being 9 the number of parameters that needs for describing three triangular shaped MFs. 
The FIS composed by 27 rules is thus characterized by a chromosome consisting of $36 +27 = 63$ elements. Consequently, the whole chromosome, defining the two FIS in the proposed FLC, will account for 126 real bounded values.
\subsection{Fuzzy Logic Controller optimization: Hierarchical Genetic Algorithm}
\label{sec:HGA}
Given an instance of a basic FLC, as the one described in Sec. \ref{sec:SFCS}, the task is now to try to optimize together with the MF parameters and weights also the number of fuzzy rules. In other words starting from the adaptive grid partition structure generated by the antecedents of the fuzzy reasoning, we want to find a minimal set of rules. The basic FLC, in fact, operates the inference with a maximal set of fuzzy rule that are encoded following a suitable scheme depicted in Fig. \ref{fig::enc_scheme}. In the search process, considering the number $K$ of fuzzy rules as a further degree of freedom, it is possible to improve the performance of the overall mechanism of inference and at the same time obtaining a simpler and more understandable structure of the fuzzy RB.
This problem can be faced, within the fuzzy-GA paradigm, using a multiple chromosome scheme. The rationale is to partition the chromosome in two separate sections $S_1$ and $S_2$, each one with a different meaning. More precisely, genes in $S_1$ can activate or deactivate genes of $S_2$. The increasing complexity of the genetic code demands suited genetic operators. By adopting a multiple chromosome scheme it is possible to decompose the problem in small sub-problems, reflecting the decomposition of the genetic code in smaller units, each one ruled by its own genetic operator. The multi-chromosome scheme, as the entire Genetic Programming discipline, is inspired by Nature. In fact, such methodology is widely used in Erokaryotes organisms, i.e. some bacteria and plant species \cite{Cavill05}. This approach, whose performances are studied in \cite{Pierrot97}, is adopted only when the problem at hand need such characteristic representation of the chromosome.  
In \cite{Pierrot97} authors apply the multiple chromosome scheme to a mixed integer programming problem consisting in the control of an industrial machinery, where the need is to activate the machinery $A$ only when it is compliant with the overall cost reduction. The multiple chromosome has a section containing the \textit{control genes} that encode the activation status of a particular machinery, while the remaining part, consisting of \textit{parametric genes}, encodes for the amount of produced goods. In \cite{Kiraly10} the multiple chromosome scheme is adopted for the solution of the multiple Traveling Salesman Problem (mTSP), that is a complex combinatorial optimization problem. \\
A particular approach to the multiple chromosome scheme is the \textit{hierarchical chromosome} one, that leads to a family of GAs, known as Hierarchical Genetic Algorithms (HGAs). The multiple chromosome representation is based on partitioning it in two sections, one containing the control genes and the other the parametric genes. The control genes encode for the activation of a particular sub-section of the parametric genes. 

\subsubsection{HGA chromosome representation}
\label{sec:MCR}
The adopted chromosome encoding scheme is based on a mixed binary and real-valued string \cite{Ko06}, where the real-valued genes encode the MF parameters and the rule weights, while the binary genes, as control genes, encode the activation of MFs. 
\begin{figure}[htbph]
\centering
\includegraphics[scale=0.57]{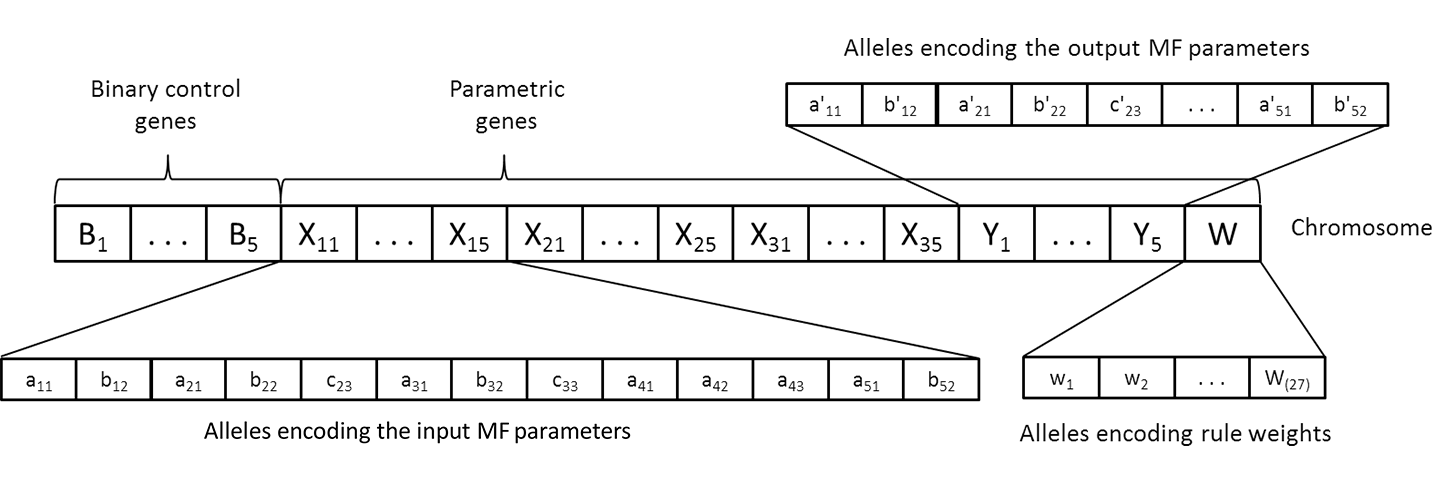}
\caption{The HGA chromosome scheme adopted in the FLC. In the first section of the chromosome we have binary genes, the \textit{control genes}, that encode for the activation or deactivation of the MFs and their respective parameters. The latter are encoded with suitable subset of the genes: the \textit{parametric genes}.}
\label{fig::Multiple_enc_scheme}      
\end{figure}
In Fig. \ref{fig::Multiple_enc_scheme}  is depicted the adopted encoding scheme for both FISs. The overall chromosome can be formally written as:
\begin{equation}
\mathbf{p}=[\mathbf{p}_{bin} \; \mathbf{p}_{real}],
\label{pbin-preal}
\end{equation}
where $\mathbf{p}_{bin}=[p_{1}, p_{2},\dots,p_{N}], \; p_{i}\in\left\lbrace0, 1\right\rbrace, \; \forall i=1,\dots,N$, being N the total number of the input MFs and $\mathbf{p}_{real}=[p_{N+1}, p_{N+2},\dots,p_{M}], \; p_{j}\in R, \; \forall j=N+1,\dots,M$, being $M$ the total number of MFs and fuzzy rules encoded in the chromosome. \\
Considering as granulation of each input 2 external trapezoidal shaped MFs, described by 2 parameters each, and 3 internal triangular shaped MFs, described by 3 parameters each, the chromosome structure for one FIS is the following.
\begin{itemize}
\item The number $N_{bin}$ of control genes is given by the product $N_{bin}=N_{MF_{in}} \times N_{in}$, where $N_{MF_{in}}$ is the number of the input MFs and $N_{in}$ is the number of the FIS input variables. Thus, for for an input described by 5 MFs, we have $3 \times 5 = 15$ control genes.
\item The number $N_{real}$ of parametric genes is composed by the number of input and output MF parameters $13+13+13+13=52$ followed by the number of rules weights, $5 \times 5 \times 5 = 125$. Thus, we have $N_{real} = 52 + 125 = 177$ parametric genes.
\item Finally, the total length of the genetic code is equal to $ 177 + 15 = 192$ genes.
\end{itemize}
Consequently, the whole chromosome, defining the two FIS in the proposed FLC, will account for 384 real bounded values. \\
\subsubsection{GA initialization and search space constraints}
\label{sec:GA_in}
Among the several ways available for generating the initial population, in the current work it is adopted the simple random generation, compliant with given suitable constraints, with the aim to cover the overall search space for the GA. 
Regarding the admissible values of the alleles encoding the MF parameters, we follow the method presented in \cite{cordon2001a}.

As concerns the constraints for the weights of the fuzzy rules,  $(w_{1},w_{2},\dots,w_{K},)$, each one is allowed to range in $[0,1]$. Once we define the constraints, the initial values for each allele can be generated through the extraction of a uniformly distributed random variable within the interval constraints.
For the control genes, as binary strings, the domain of the random variables is the binary valued set $\{0,1\}$ and the random extraction process assigns equal probability to the two possible values.
\subsubsection{Mixed genetic operators}
\label{sec:Mixed_g_o}
The chromosome representation given in (\ref{pbin-preal}) is based on the distinction between the control genes and the parametric genes. This separation leads to an heavy simplification of the design process for the genetic operators. It is well-known that the crossover operator is an important genetic operator that can effectively exchange and recombine genetic material from two selected parents. The hierarchical scheme allows to a semantic separation of the genetic material leading to design a mixed crossover operator consisting of a specific simple operator for control genes and a suitable operator for the parametric genes. Hence, the adopted mixed crossover is realized as follows:  
\begin{itemize}
\item for the control genes $\mathbf{p}_{bin}$ it is designed a simple ``one cut-point'' crossover where the cutting point is set by randomly drawing from a uniform distribution;
\item for the parametric genes $\mathbf{p}_{real}$ it is applied a ``convex crossover operator" described by the following relations \cite{Ko06}:
\begin{eqnarray}
p_{new1}=\lambda\: p_{real1}+(1-\lambda)\: p_{real2} \label{eq:xover-conv1}\\
p_{new2}=\lambda\: p_{real2}+(1-\lambda)\: p_{real1},
\label{eq:xover-conv2}
\end{eqnarray}
where $p_{new1}$ and $p_{new2}$ are the offspring's chromosomes, while $p_{real1}$ and $p_{real2}$ are the parent's chromosomes and $\lambda$ is a random value, such as $\lambda\in[0,1]$.
\end{itemize}
The mutation operator is crucial for the exploration capability of the overall optimization algorithm, allowing a suitable variability in the population. More specifically, the mutation operator is designed to change one or more genes depending on a suitable mutation rate used as threshold for a given probability value for the mutation of each allele. Even for the mutation operator it is adopted a mixed scheme:\\
\begin{itemize}
 \item for the binary section $\mathbf{p}_{bin}$ of the chromosome it is applied a single point mutation operator in which the mutation point is chosen randomly.\\
\item  for the real-valued section $\mathbf{p}_{real}$ of the chromosome the ``non-uniform" mutation operator is performed. The genetic operator takes into account the age of the given population in a way that the mutation effect vanishes with the progression of the generations. More specifically, given a chromosome $\mathbf{p}_{real}$ for a selected gene  $p_{real,k}$, the resulting offspring gene is generated choosing randomly among the following relations \cite{Ko06}:\\
\begin{small}
\begin{eqnarray} 
p_{new,k}=p_{real,k}+\Bigl(p_{real,k}^{U}-p_{real,k}\Bigr) b\Biggl(1-\frac{g}{G}\Biggr)^a  \label{eq:mut-non-unif1} \\
p_{new,k}=p_{real,k}+\Bigl(p_{real,k}-p_{real,k}^{L}\Bigr) b\Biggl(1-\frac{g}{G}\Biggr)^a, \label{eq:mut-non-unif2}
\label{eq:non_uniform_mut}
\end{eqnarray}
\end{small}
\noindent where $p_{real,k}^{U}$ and $p_{real,k}^{L}$ are, respectively, the upper and lower bounds for $p_{real,k}$, $b\in[0,1]$ is a random uniformly distributed real value, \emph{g} is the current generation index and \emph{G} the predefined maximum number of generations. Finally, $a$ is a parameter determining the decay speed. 
\end{itemize}
We remark that the main difference between the two evolutionary approaches, namely classic fuzzy-GA and fuzzy-HGA, lies in a higher number of degrees of freedom available for the FIS optimization. In fact, a composite encoding scheme capable of turning off/on each fuzzy set allows a more effective exploration of the search space. If we call $\mathbb{S}$ the set of all FIS$\left(\Theta\right)$ depending on a set of structural parameters $\Theta=\{\theta_{MF},\theta_w,\theta_{r}\}$, where the subscript $MF,w,r$ refer to the membership function parameters, the rule weights and the fuzzy rules, respectively, the fuzzy-HGA scheme is able to explore a subset $\mathbb{S}^{'} \subseteq \mathbb{S}$. Borrowing the Linear Algebra jargon, the fuzzy-HGA scheme can explore a subspace of the overall space which hosts the FISs. $\mathbb{S}$ can contain FISs with poorly performing fuzzy rules that the GA could evaluate in order to optimize the objective function, lowering in this way the overall search efficiency. 
In fact, among a huge set of fuzzy rules, many of them can be useless (for example because are centered onto input space regions corresponding to infrequent system states), or even harmful for the overall performances (for example, if the parameters of the related membership functions are not well tuned). Thus, according to the Occam’s Razor criterion, a tuning procedure able to perform a reduction of the structural complexity of the Rule Base (and thus of the FIS) is fundamental to improve the overall performances of the EMS

\section{Experimental Evaluation}
\label{sec:Experim_ev}
This section reports several experimental evaluations with the aim to measure the performance of the fuzzy-HGA control system adopted for the energy flow optimization task in the proposed MG model. Specifically the performance of the proposed FLC is measured in a suitable scenario on a time-span of one year reporting the results for several configurations of the energy storage device connected to the MG and for two mutation parameters of the given GA. 
Finally, it is provided a comparison with the basic fuzzy-GA paradigm described in \ref{sec:Simple_scheme} and studied in our previous work \cite{de2013genetic}, where only the MF parameters and the rule weights are tuned. 

\subsection{Experimental setting}
\label{sec:Scenery}

\begin{figure}[htbph]
\centering
\includegraphics[scale=0.85]{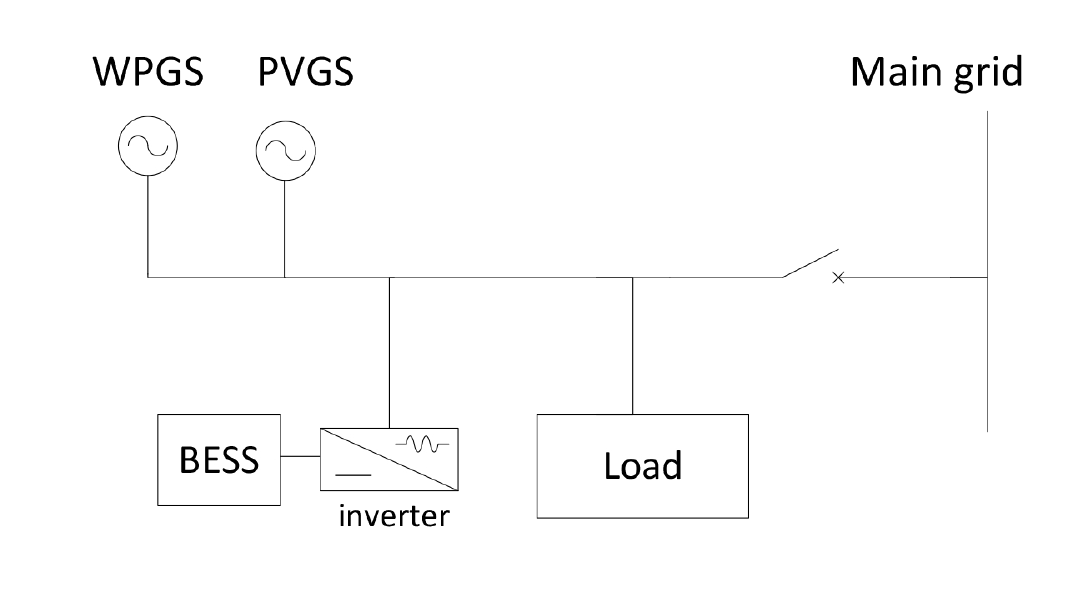}
\caption{Scheme of the considered MG.}
\label{fig::MG_scheme}      
\end{figure}

The scenario considered for experiments consists of a medium dimension MG. The basic scheme is depicted in Fig. \ref{fig::MG_scheme}. The  MG  model is equipped with a distributed resources like a photo-voltaic generator (PVGS) and a wind generator (WPGS). The model is completed with an energy storage capacity of 24 kWh and the control system (EMS) based on fuzzy-HGA paradigm where a hierarchical evolutionary algorithm (HGA) is in charge of learning the FLC parameters and, at the same time, the number of adopted fuzzy rules. The control system manages the energy flow of the MG in input and output maximizing the accounting profit, dispatching control actions depending on: the energy balance, the SOC of the battery and the price signal of the energy -- see Sec. \ref{sec:PF}. 

%

\begin{figure}[ht!] 
\centering
\includegraphics[scale=0.6]{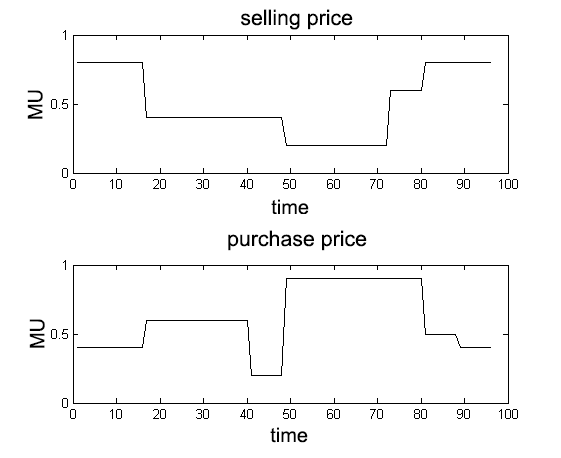}
\caption{Time-profile of the Energy price in selling and purchasing.}
\label{fig:cost-prof}
\end{figure}

Experiments are performed on real-world data set provided by Acea Distribuzione S.p.A., the electric distribution utility that manages the power grid of Rome in Italy. The electricity demand and production time profiles regard measurements in a time-span of one year and are sampled every 15 minutes. The overall dataset has been split in two disjoint set, \textit{training set} and \textit{test set}, considering the odd and the even time samples, respectively. The parameters of the FLC are learned on the training set and the performances of the EMS are evaluated on the test set. The daily cost profiles, in generic monetary units (MU), are reported in Fig. \ref{fig:cost-prof}. It is noted that the adopted electricity prices are multi-hours tariffs. In fact, the selling price is high during the day when the demanded energy is high, while it is lower in the night when the power is generated by only the wind source.

\begin{figure}[ht!] 
\centering
\includegraphics[scale=0.6]{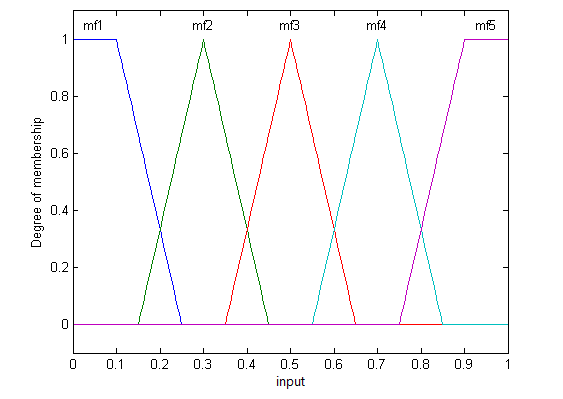}
\caption{Adopted Membership Functions for both FISs.}
\label{fig:fisalfa-orig}
\end{figure}

The model of the energy storage capacity is the one described in Sec. \ref{sec:BM} referring to the circuital scheme of Fig. \ref{fig:batt_scheme}.
The storage unit used for experiments is modeled on the basis on a SCiB battery, developed by Toshiba, with storage capacity of 24 kWh. Simulations are performed on the basis of two main configurations resumed in Tab. \ref{tab:config-batt}. The defuzzification method used is the Mean of Maximum (MoM).
\begin{table}[ht!] 
\centering
\scalebox{0.9}{
\begin{tabular}{ccc}
\toprule
& Configuration 1 & Configuration 2 \\
\midrule
$R_{int} \quad (m \Omega)$ & 2 & 1.5 \\
$I_{bat} \quad (A)$ & 8 C & 8 C \\
$Q_{bat} \quad (Ah)$ & 80 & 80 \\
$\eta_{bat}$ & 0.86 & 0.9 \\
$SOC_{ini} \quad (\%)$ & 40 & 80 \\
$SOC_{min} \quad (\%)$ & 0 & 15 \\
\bottomrule
\end{tabular}}
\caption{Parameters defining the two evaluated configurations for the Toshiba SCiB battery model.}
\label{tab:config-batt}
\end{table}
For the two FISs, the starting configuration is depicted in Fig. \ref{fig:fisalfa-orig}.
\begin{figure}[ht!] 
\centering
{\includegraphics[scale=0.65]{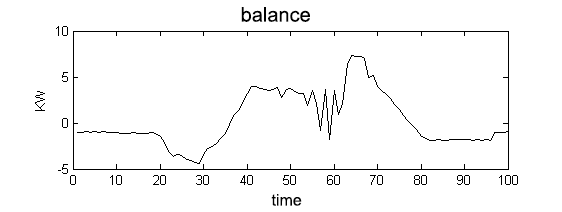}}
\caption{The time-profile of the power balance in input to the system, ranging in one day.}
\label{fig:balance_profile}
\end{figure}

%
The fuzzy rules optimization task is obtained by a Hierarchical Genetic Algorithm (HGA) as described in Sec. \ref{sec:HGA}. The initial population is composed by 40 individuals. 
The genetic operators used for the proposed HGA are described in Sec. \ref{sec:Mixed_g_o}. 
The selection operator is the ``stochastic uniform", while the crossover and the mutation operators are of mixed type, such as the ``one point" crossover and mutation adopted for the control genes, while for the parametric genes a ``convex" crossover (\ref{eq:xover-conv2})  and a ``non-uniform" mutation (\ref{eq:non_uniform_mut}) are used. 
The decay speed $a$ is set to 1, the ``crosover fraction" is 0.8, while the mutation rate $\rho$ is evaluated for two values: 0.001 and 0.01. Finally the  proposed objective function is given by expression (\ref{eq:Obj_f}).

\subsection{Preliminary experiments}
\label{sec:Preliminary_Tests}
\begin{figure}[ht!] 
\centering
\includegraphics[scale=0.75]{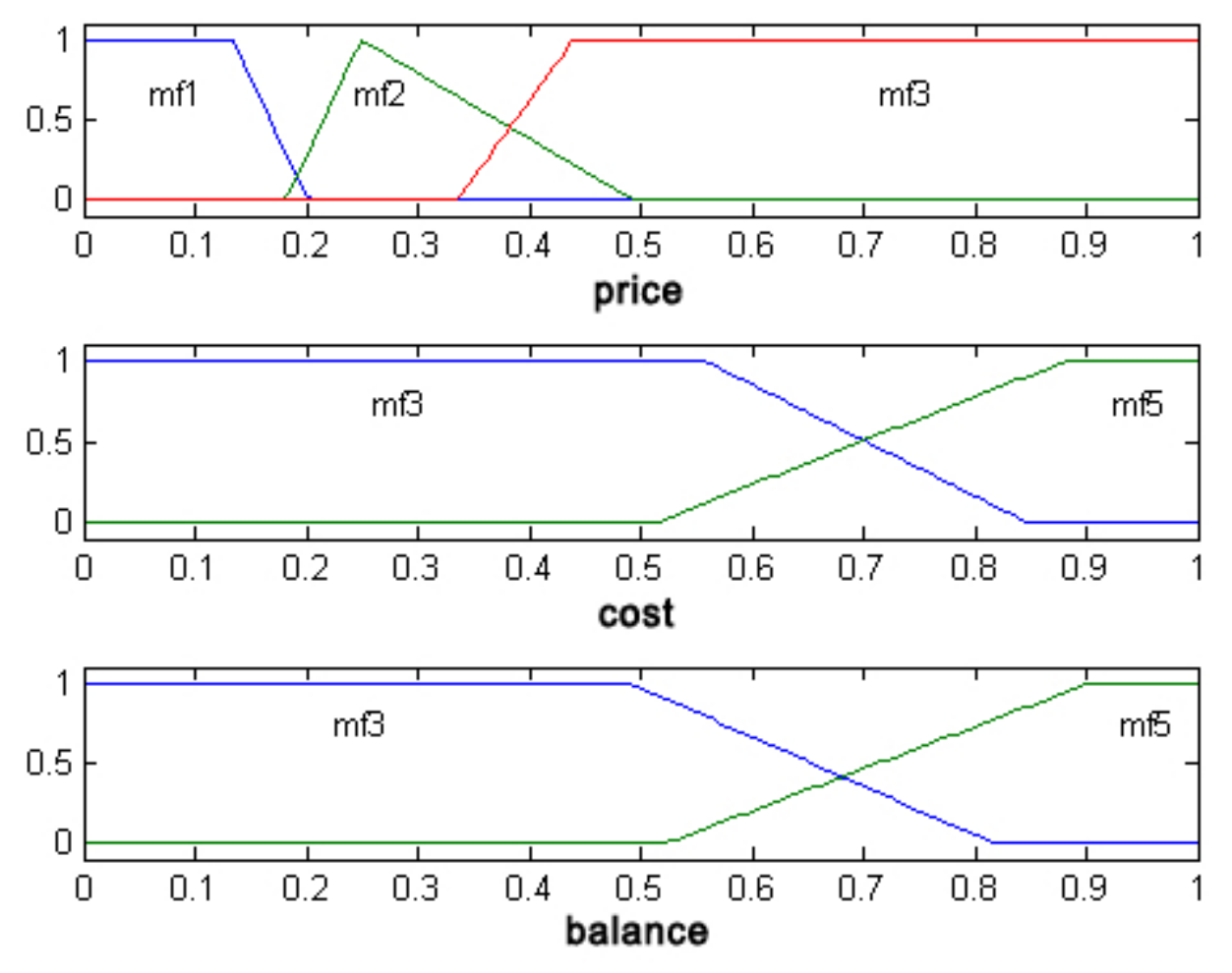}
\caption{Optimized Membership Functions for the FIS computing the $\alpha$ parameter for Config. N° 1.}
\label{fig:fisalfa-MF-ott}
\end{figure}

\begin{figure}[ht!] 
\centering
\includegraphics[scale=0.75]{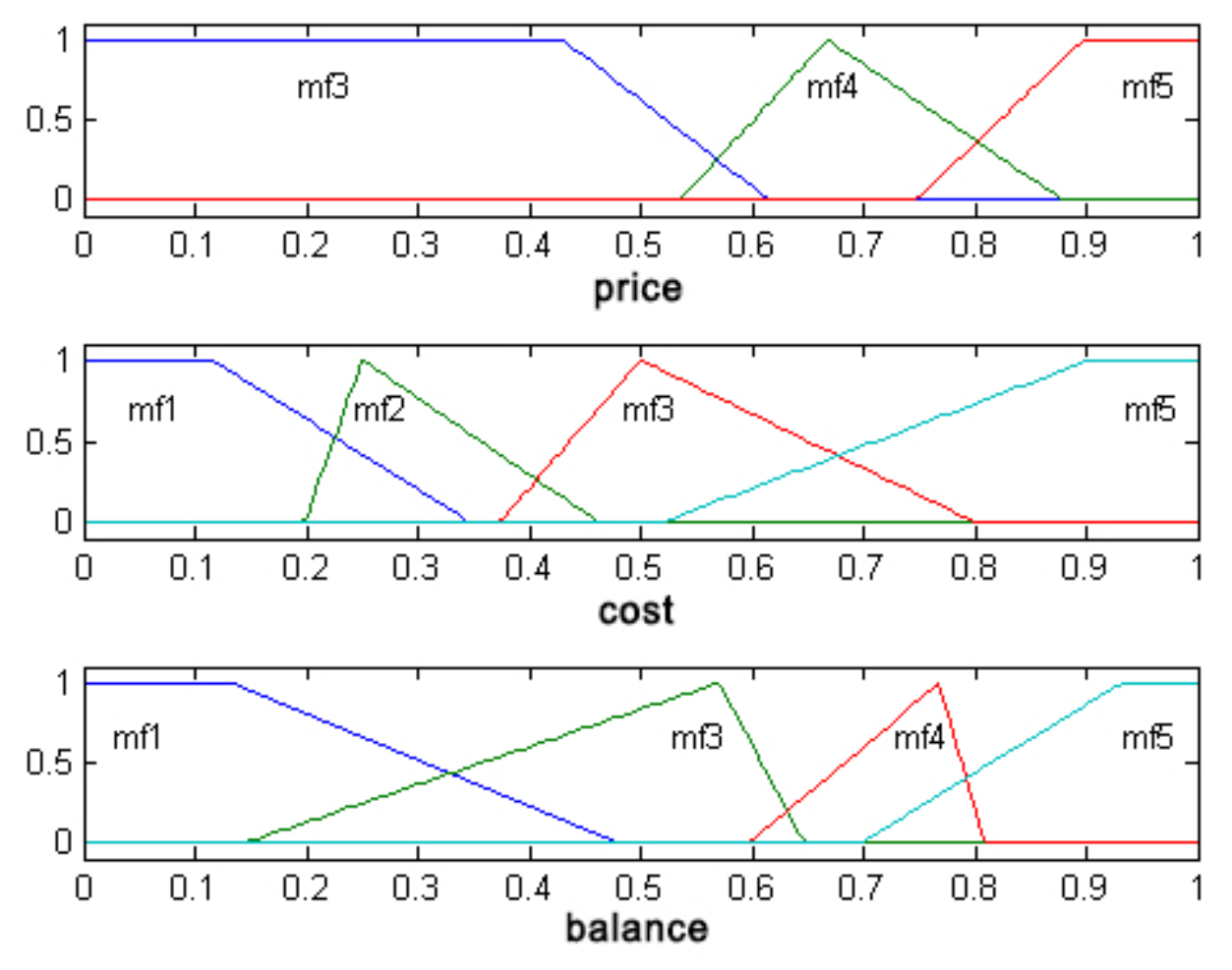}
\caption{Optimized Membership Functions for the FIS computing the $\beta$ parameter for Config. N° 1.}
\label{fig:fisbeta-MF-ott}
\end{figure}

\begin{figure}[ht!] 
\centering
\includegraphics[scale=0.7]{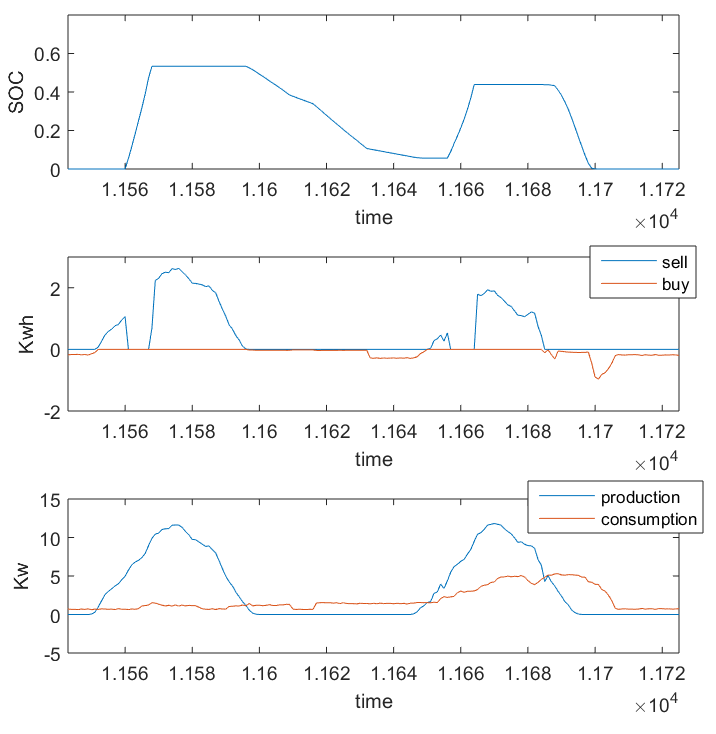}
\caption{Configuration 2: SOC time profile (top), time profiles of energy purchased from and sold to the main-grid, power production and demand time profiles (bottom).}
\label{fig:test2-perf}
\end{figure}

\begin{figure}[ht!] 
\centering
\includegraphics[scale=0.7]{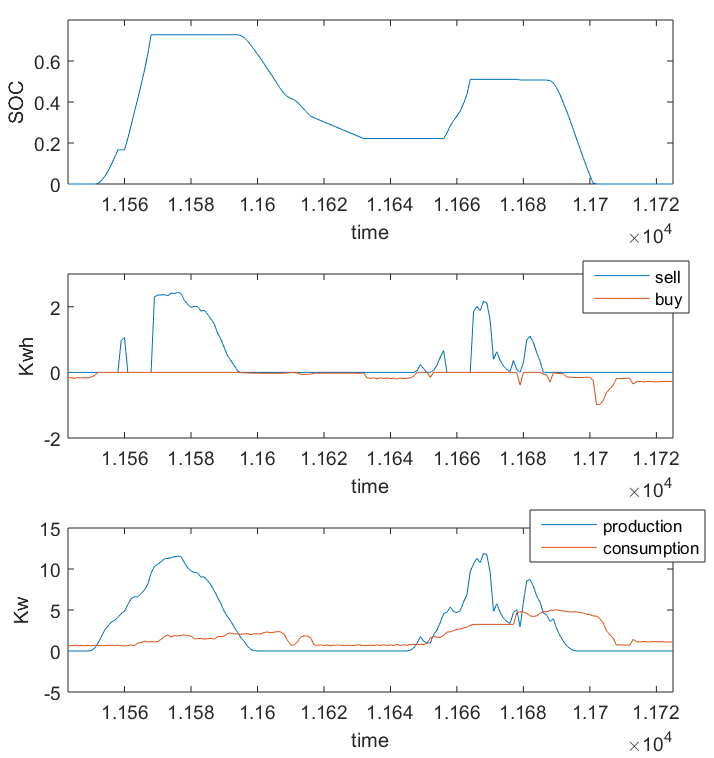}
\caption{Configuration 3: SOC time profile (top), time profiles of energy purchased from and sold to the main-grid, power production and demand time profiles(bottom).}
\label{fig:test3-perf}
\end{figure}

\begin{figure}[ht!] 
\centering
{\includegraphics[scale=0.7]{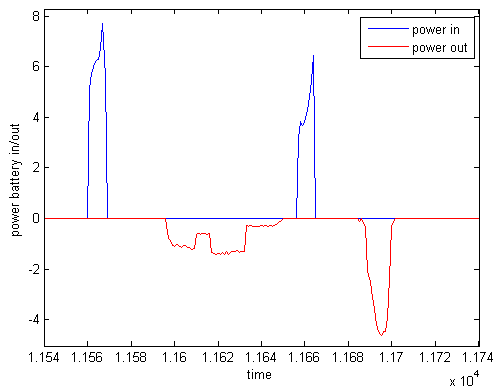}}
\caption{Battery power input and output.}
\label{fig:test23-power-batt}
\end{figure}

\begin{figure}[ht!] 
\centering
{\includegraphics[scale=0.75]{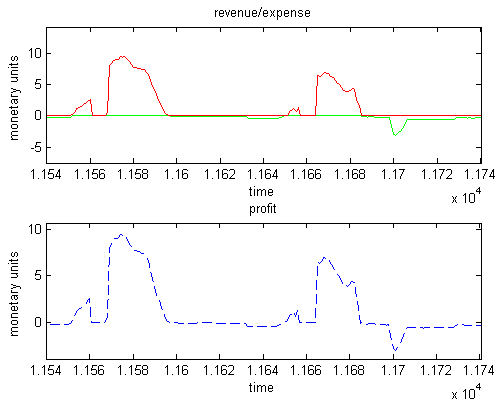}}
\caption{Revenues and expenses (top) and profit (bottom).}
\label{fig:test23-revexp}
\end{figure}

As preliminary experiments we compare the two battery configurations resumed in Tab. \ref{tab:config-batt} with two scenarios characterized by two different mutation rate values: $\rho=\{0.01, 0.1\}$. The electricity prices, as input data, are those described above in Sec. \ref{sec:Scenery} and range in the entire year. 
The HGA searching results within the Config. N° 1 and $\rho=0.01$, leaded to the MFs depicted in Figs. \ref{fig:fisalfa-MF-ott}, \ref{fig:fisbeta-MF-ott} with the following control chromosome sections -- see Sec. \ref{sec:MCR}:
\begin{eqnarray}
p_{bin}(\alpha)=[1\;1\;1\;0\;0\;|\;0\;0\;1\;0\;1\;|\;0\;0\;1\;0\;1]; \\
p_{bin}(\beta)=[0\;0\;1\;1\;1\;|\;1\;1\;1\;0\;1\;|\;1\;0\;1\;1\;1].
\end{eqnarray}
Results reported in Tab. \ref{tab:annual-test} show how the FLC obtains good accounting profit performances also using an higher mutation rate. It is noted how the Config. N° 2, characterized by a  higher efficiency and a lower threshold limit ($SOC_{min}=15\%$) for the SOC, leads to a further improvement of the accounting profit due to the lower expenses related to energy purchase. The main causes can be found in: $i)$ a better use of the energy storage system with an high value of the efficiency that allows to store a greater amount of electric charge leading to lower losses; $ii)$ the introduced threshold value constrains the controller to undertake ``crisp actions" charging the battery when the SOC is lower than $SOC_{min}$. The above described behavior can be observed in Figs. \ref{fig:test2-perf}, \ref{fig:test3-perf} where are reported, for a suitable time interval: the SOC, the power production, the power consumption, the energy purchased and the energy sold for the second and third configuration, respectively, and reported in Tab. \ref{tab:annual-test}. Fig. \ref{fig:test23-power-batt} shows, for the same configurations, the electric power drawn from and stored in the battery, while in Fig. \ref{fig:test23-revexp} are depicted the expenses, the revenues and the accounting profit time profiles of one sample day. As concerns the number of rules returned by the fuzzy-HGA algorithm for all the configurations evaluated, it has been found lower than the maximum value corresponding to the complete coverage of the input space using all the available MFs.

%
%
%
\begin{table}[ht!] 
\centering
\scalebox{0.9}{
\begin{tabular}{ccccc}
\toprule
& Config. 1 & Config. 1 & Config. 2 & Config. 2 \\
& $\rho=0.1$ & $\rho=0.01$ & $\rho=0.1$ & $\rho=0.01$ \\
\midrule
Expense (MU) & 9379.498 & 9790.907 & 8628.261 & 8545.162  \\
Revenue (MU) & 13371.530 & 13864.880 & 12880.630 & 12822.870  \\
Profit (MU) & 3992.030 & 4073.976 & 4252.373 & 4277.713  \\
\# FIS rules ($\alpha$) & 57 & 72 & 49 & 61  \\
\# FIS rules ($\beta$) & 65 & 49 & 72 & 57  \\
\bottomrule
\end{tabular}}
\caption{Accounting performance and number of fuzzy rules retrieved for both FISs $\alpha$ and $\beta$,  obtained on the test set for different system's configuration.}
\label{tab:annual-test}
\end{table}
\subsection{Comparison with the classic fuzzy-GA method}
\label{sec:Comparison}
The following section proposes a performance comparison between the herein studied fuzzy-HGA scheme and the classic fuzzy-GA scheme reported in \cite{de2013genetic} and resumed in Sec. \ref{sec:Simple_scheme}.  
As concerns the fuzzy-HGA scheme, we have the same considerations reported in Sec. \ref{sec:Scenery}. Both systems are provided with same input data and parameters as concerns the MG model. The number of GA generations is fixed to the value of 100 and the initial population, initialized randomly, consists of 40 individuals. The fitness function is the one reported in (\ref{eq:Obj_f}). For the configuration of the energy storage system both MGs are provided with an accumulator with 24 kWh capacity whose configuration is the N° 2 ($\rho=0.01$) reported in Tab. \ref{tab:config-batt}.
\begin{table}[ht!] 
\centering
\scalebox{0.9}{
\begin{tabular}{ccc}
\toprule
& Fuzzy-GA System & Fuzzy-HGA System \\
\midrule
Expense (MU) & 11066.350 & 8545.162 \\
Revenue (MU) & 13626.800 & 12822.870 \\
Profit (MU) & 2560.446 & 4277.713 \\
\bottomrule
\end{tabular}}
\caption{Financial accounting performances obtained on the test set.}
\label{tab:test-old-3}
\end{table}
%

\begin{figure}[ht!] 
\centering
{\includegraphics[scale=0.73]{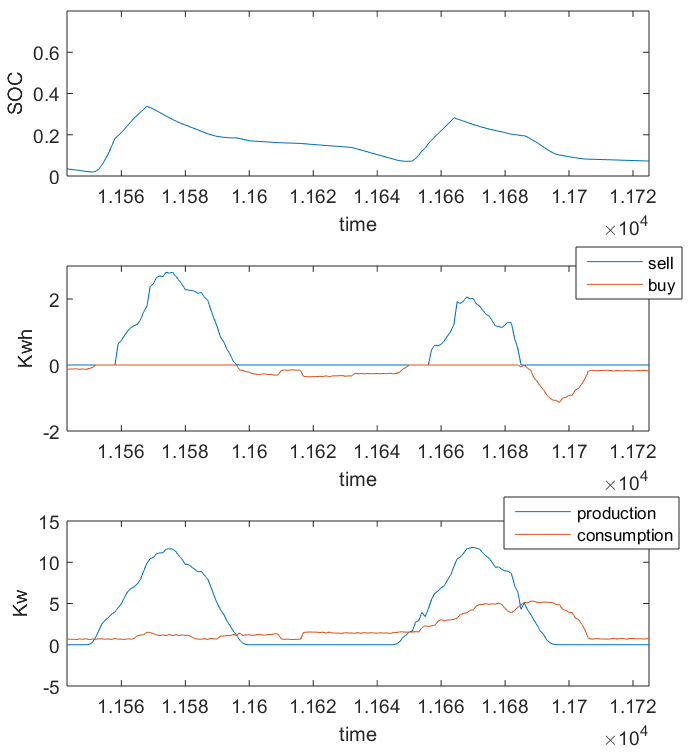}}
\caption{Economic performances of the fuzzy-GA optimization schemes: SOC time profile (top), electricity purchased from and sold to main-grid time profiles electricity production and demand time profiles (bottom).}
\label{fig:test-old-3-perf}
\end{figure}

\begin{figure}[ht!] 
\centering
\subfloat[][Fuzzy-GA System]
{\includegraphics[scale=0.73]{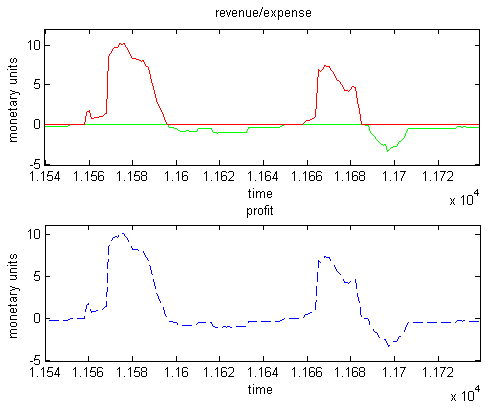}}

\subfloat[][Fuzzy-HGA System]
{\includegraphics[scale=0.73]{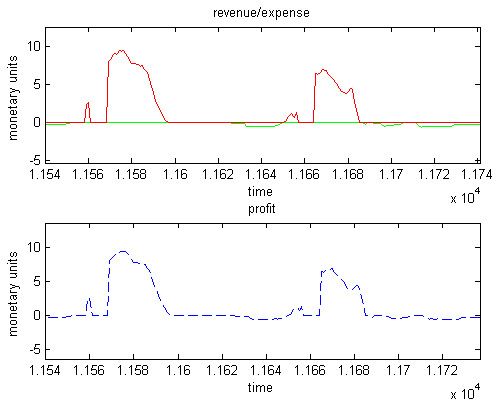}}
\caption{Revenues and expenses time profiles (top) and profit (bottom).}
\label{fig:test-old-3-revexp}
\end{figure}

Figs. \ref{fig:test-old-3-perf} and \ref{fig:test-old-3-revexp} show the time profiles of the battery SOC together with the power produced and demanded, the energy sold and purchased and the overall financial accounting performances. 


%

%
Results in Tab. \ref{tab:test-old-3} show an increase of performance for the fuzzy-HGA system compared with the classic fuzzy-GA approach. In fact the hierarchical structure of the GA allows to learn, in addition to the shape of MFs and the rules weights, also the number of rules searching for a suitable cardinality of the RB. This greater degree of freedom is obtained allowing more adaptation capability of the FIS to the environment with the cost of a greater complexity of the whole GA algorithm. 
However, this increment in the learning procedure complexity has the advantage of reducing the structural complexity of the optimized FIS, in terms of both the number of rules and the number of antecedents, as shown for instance in Fig. \ref{fig:fisalfa-MF-ott} and Fig. \ref{fig:fisbeta-MF-ott}. This results in simpler FISs, lowering the overall computational cost of the EMS when taking decisions in real time.

\section{Conclusions}
\label{sec:conclusions}
Computational intelligence techniques are today a consolidated framework for solving engineering problems such as challenges arising in Smart Grid context. In this paper we study a portion of a Smart Grid, acting as a microgrid, characterized by the presence of renewable sources and equipped with a BESS unit allowing trading operations with the main-grid related to energy exchanges. The considered MG model has been sized for small-scale applications, such as small energy grids in rural areas or small housing units in urban areas, considered as atomic elements of a larger a multi-microgrid system.
The decision making task is carried on through a Fuzzy Logic Controller optimized by a Hierarchical Genetic Algorithm able to tune FLCs parameters (i.e. the MF parameters and the fuzzy rule weights) minimizing, at the same time, the number of fuzzy rules. The fuzzy rules number optimization is carried out in FLCs thanks to the hierarchical structure of the adopted chromosome representation consisting of two main sections having a different semantic meaning. The control section, encoded as a binary string, controls the activation of the parametric section representing the MF parameters and the weights of fuzzy rules. 
The specific encoding scheme of the HGA offers a higher flexibility in tuning the FLC in comparison with the classic scheme which foresees a fixed and maximal number of rules due to the rigidity of the chromosome length. 
The proposed method for EMS synthesis incorporates a learning step. In fact, once FIS rules are optimized by considering training set data through an off-line procedure running on a plain workstation, the trained rules can be uploaded into an embedded system (a microcontroller) ready to work in real-time. From this point of view, the added complexity of the HGA scheme can be well managed off-line by an inexpensive workstation (such as an Intel i7 6thh generation with 32 Gb of ram), exploiting its power in evolving suitable rules without affecting the real time operations. Note that the final aim of the HGA algorithm is the reduction of the structural complexity of the rule base in the FIS, dropping the number of rules and the number of antecedents in each rule. Consequently, with respect to a FIS resulting from a learning procedure based on a plain version of a Genetic Algorithm, the microcontroller is in charge to handle a lower number of rules in performing decision making, in a given time interval. Moreover, note that the typical dynamics characterizing a microgrid (at the considered abstraction level) are slow enough to set the length of the time interval in the order of minutes (15 minutes, in the considered case). Consequently, even an entry level and inexpensive microcontroller (such as an “Arduino Due” board, based on the Atmel SAM3X8E ARM Cortex-M3 CPU) has sufficient computational power to handle a great number of fuzzy rules in such time intervals.
The fuzzy-HGA paradigm is thereby evaluated for some configuration of the MG parameters, battery model parameters and genetic operator parameters with the aim to study the overall behavior of the fuzzy control system. A further comparison is performed between a classic optimization scheme within the fuzzy-GA paradigm and the adopted fuzzy-HGA. The classic GA scheme foresees the optimization of the MF parameters together with the fuzzy rule weights in a fixed Rule Base (RB) whose cardinality is maximal (grid partition). The fuzzy-HGA algorithm outperforms the classic fuzzy-GA scheme by 67\%. The hierarchical encoding method of the FIS parameters and the definition of suited genetic operators explain the improvement in performance reached at the cost of a more complex design of the GA and of the encoding scheme. However, this additional computational cost at design and training stages is balanced with the possibility to obtain simpler FLCs in terms of number of rules, allowing the real time implementation of the complete control system on low cost embedded electronic devices. 

\section{Future works}
\label{sec:FWS}
Results are encouraging and suggest to work in future developments on both MG modeling and learning. 
Future developments foresee the definition of suitable measures, such as i) the stress of the battery in charging and discharging processes, ii) the stress of the main-grid in terms of undesired fast changes of power flow between the MG and the main-grid. These measures can be considered as additional objectives (that are likely conflicting with the main one based on prosumer profit) aiming at controlling both the stress of the storage system and that of the main-grid. A mono-objective or a muti-objective GA scheme can be adopted taking into account the possibility of tuning the fuzzy controller allowing different behaviors, such as promoting the auto-consumption, lowering the battery stress  or the workload on the main-grid.
Furthermore, an interesting future scenario consists in performing a spatial granulation of the whole considered distribution network, where each territorial granule acts as a MG, each one able to exchange energy with other spatially close MGs and/or with the main-grid, with different pricing programs. The MGs can have different characteristics in terms of aggregate demand, production, energy source type and storage systems. Moreover, to increase the model reliability, it is interesting to consider power losses in energy exchange between MGs. The intrinsic heterogeneity of MGs together with the loss factors and the learning features fulfilled within the fuzzy-HGA paradigm allow to study the behavior of the MG acting as agents, with possible emerging cooperative behavior in satisfying objective functions defined by a higher level controller. Hence, in this augmented scenario some of the initial assumption adopted through the model can be relaxed and the MGs can be equipped with a suitable model of telecommunication network, taking into account also the capacity constraints in data delivery. Moreover the MG can foresee a demand composed of schedulable loads such as smart appliances and electric vehicles, allowing to design suitable demand response (DR) programs able to act together with the FLC aiming to optimize predefined objectives. From the learning side, it is really interesting to apply other nature inspired methods such as autoimmune or bacterial optimization algorithms for the optimization of the FLCs. Furthermore, a FLC with a Sugeno inference core can be adopted as the main fuzzy decision module and the HGA could be suitably adapted in order to obtain a fair comparison with the Mamdani controller.%

\bibliographystyle{abbrvnat}

\end{document}